\documentclass{article}

\usepackage{arxiv}

\usepackage[utf8]{inputenc} 
\usepackage[T1]{fontenc}    
\usepackage{hyperref}       
\usepackage{xcolor}
\hypersetup{
    colorlinks,
    linkcolor={red!50!black},
    citecolor={blue!50!black},
    urlcolor={blue!75!black}
}
\usepackage{url}            
\usepackage{booktabs}       
\usepackage{amsfonts}       
\usepackage{nicefrac}       
\usepackage{microtype}      
\usepackage{lipsum}		
\usepackage{graphicx}
\usepackage{natbib}
\usepackage{doi}

\usepackage[capitalize]{cleveref}
\Crefname{figure}{Figure}{Figures}
\Crefname{tabular}{Table}{Tables}
\Crefname{section}{Section}{Sections}

\title{Data-Centric AI in the Age of Large Language Models}

\date{} 					

\author{
    Xinyi Xu$^{1,2}$\thanks{
    $^1$National University of Singapore, 
    $^2$Agency for Science, Technology and Research, 
    $^3$Singapore-MIT Alliance for Research and Technology Centre,
    $^4$Guangdong Lab of AI and Digital Economy~(SZ), 
    $^5$CNRS@CREATE,
    $^6$Massachusetts Institute of Technology, 
    $^7$Allen Institute for AI,
    $^8$University of Washington.
    Correspondence to: Xinyi Xu <\textrm{xinyi.xu@u.nus.edu}>.
    } \And
    Zhaoxuan Wu$^{1,3}$ \And
    Qiao Rui$^1$ \And
    Arun Verma$^1$ \And
    Yao Shu$^4$ \And
    Jingtan Wang$^{1,2}$ \AND
    Xinyuan Niu$^{1,2}$ \And
    Zhenfeng He$^1$ \And
    Jiangwei Chen$^{1,2}$ \And
    Zijian Zhou$^{1,3}$ \And
    Gregory Kang Ruey Lau$^{1,5}$ \And
    Hieu Dao$^1$ \And
    Lucas Agussurja$^1$ \And
    Rachael Hwee Ling Sim$^1$ \And
    Xiaoqiang Lin$^1$ \And
    Wenyang Hu$^1$ \AND
    Zhongxiang Dai$^6$ \And
    Pang Wei Koh$^{7,8}$ \And
    Bryan Kian Hsiang Low$^1$ \And
}

\renewcommand{\headeright}{}
\renewcommand{\undertitle}{}
\renewcommand{\shorttitle}{Data-Centric AI in the Age of Large Language Models}

\hypersetup{
pdftitle={Data-Centric AI in the Age of Large Language Models},
pdfsubject={AI, ML},
pdfauthor={Xu et al.},
pdfkeywords={Data-centric AI, LLMs},
}

\begin{document}
\maketitle
\begin{abstract}
This position paper proposes a data-centric viewpoint of AI research, focusing on large language models~(LLMs).
We start by making a key observation that \emph{data} is instrumental in the developmental (e.g., pretraining and fine-tuning) and inferential stages (e.g., in-context learning) of LLMs, 
and yet it receives disproportionally low attention from the research community. We identify four specific scenarios centered around data, covering data-centric benchmarks and data curation, data attribution, knowledge transfer, and inference contextualization.
In each scenario, we underscore the importance of data, highlight promising research directions, and articulate the potential impacts on the research community and, where applicable, the society as a whole. 
For instance, we advocate for a suite of data-centric benchmarks tailored to the scale and complexity of data for LLMs.
These benchmarks can be used to develop new data curation methods and document research efforts and results, which can help promote openness and transparency in AI and LLM research.
\end{abstract}


\section{Introduction} \label{sec:introduction}

The latest large language models~(LLMs)~\cite{alayrac2022flamingo,anil2023palm,arXiv23_openai2023gpt,radford2021learning,ramesh2021zero,ramesh2022hierarchical,rombach2022high,touvron2023llama} are typically trained on extensive corpora of raw data scrapped from the Internet and then fine-tuned on specialized domain data.
These LLMs have demonstrated not only incredible performance on benchmarks~\citep{lee2023holistic,liang2023holistic}, but also remarkable abilities to follow and execute human instructions~\citep{ouyang2022training,wang2022supernatrual}, and to learn ``in-context''~\citep{dong2023in-context-survey} from the contextual data given by the user along with the query.
At the core of these impressive achievements, we identify that data, in different forms, scales, and usages, is a common denominator.

However, the bulk of research to date has focused on modeling improvements, and little is known about how to best use data for the developmental stages (i.e., pretraining and fine-tuning) and the inferential stage (using LLMs for inference or generation).
For pretraining, the exact composition of pretraining datasets used by many leading foundation models is proprietary~\citep{,arXiv23_google2023gemini,arXiv_chen2021evaluating_codex,alphacode2022,arXiv23_openai2023gpt}, while data scrapped from the Internet is often noisy and can pose legal and security risks~\citep{barrett2023identifying,carlini2023poisoning,henderson2023foundation,min2024silo}. Moreover, since pretraining large models is expensive (e.g., GPT-$4$ costs over \$$100$ million to build~\citep{GPT4_100million}), it is prohibitively costly to evaluate different choices of pretraining data.
These characteristics raise the difficulties of identifying the factors that underlie an effective pretraining dataset.
Then, for fine-tuning, compared to the array of modeling techniques~\citep{zhang2023instruction_tuning_survey}, the methods for data curation are under-explored~\citep{chen2024automated_data_curation_FT} and most prior works adopt manual approaches~\citep{honovich_2023_ft_by_hand,wang_2023_self_instruct_ft_by_hand,wei2022finetuned,ye2021_crossfit_ft_by_hand} which are difficult to generalize and costly to deploy at scale.

It is yet unclear what a generalizable and cost-effective approach is to design pretraining and fine-tuning data in order to push the LLMs' limits beyond what is achievable solely by better modeling techniques.

Separately, for the inferential stage, there are model-centric efforts ``optimizing the instructions'' for LLMs to improve how they utilize the user-provided contextual data~\citep{peng2023instruction,wang2023instructretro} but only relatively limited data-centric research on improving the user-supplied contextual data itself, even though the LLM's performance is shown to be sensitive to the contextual data's quality~\citep{liu2023prompt} and ordering~\citep{liu2023lost,lu2022fantastically}.

\begin{figure}[!ht]
    \centering
    \includegraphics[width=0.7\linewidth]{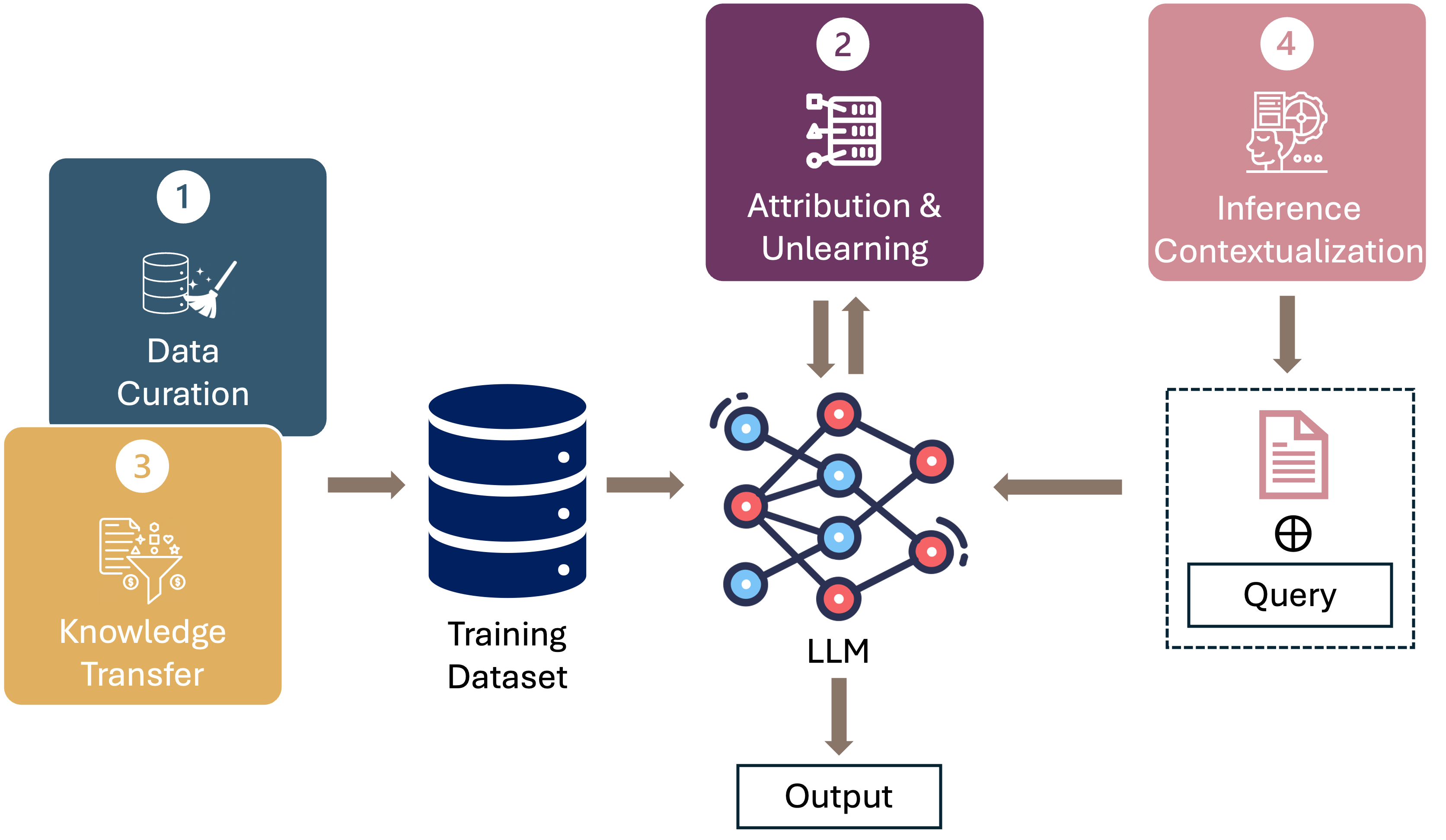}
    \caption{
    \cref{sec:benchmarking} (indexed 1 in the figure) underscores the importance of the training data (for both pretraining and fine-tuning) and the data curation techniques. 
    \cref{sec:attribution} (indexed 2 in the figure) highlights that the  LLMs' outputs depend on the training data.
    \cref{sec:distillation} (indexed 3 in the figure) describes the ``knowledge'' of the LLMs to be transferred from some training data.
    \cref{sec:context} (indexed 4 in the figure) demonstrates the usage of data by the LLMs at inference (i.e., response to a query).
    }
    \label{fig:overview}
\end{figure}

We advocate for data-centric research that can turn the art of using data into science and unlock the next generation of more effective and compact LLMs.
Our position is framed within the following four scenarios of different interactions between LLMs and data; refer to \cref{fig:overview} for a diagrammatic overview.
For each scenario, we highlight the unique characteristics and challenges, identify motivating use cases and promising research directions, and discuss potential impacts.
We do not claim to be the first to propose these directions, but rather aim to underscore the importance of the data-centric perspective and its impacts.
While our exposition is not exhaustive, we hope our ``first cut'' at a holistic viewpoint of data-centric research can generate more discussion and inspire innovation.

\paragraph{Benchmarks and curation for training data.} The recent successes of LLMs such as ChatGPT~\cite{arXiv23_openai2023gpt}, PALM~2~\cite{anil2023palm}, and LLaMA~2~\cite{touvron2023llama}, as well as vision-language models including CLIP~\cite{radford2021learning}, Flamingo~\cite{alayrac2022flamingo}, Stable Diffusion~\cite{rombach2022high} and DALL-E~\cite{ramesh2021zero,ramesh2022hierarchical}, are powered by large, heterogeneous datasets rather than solely by advanced modeling techniques.
CLIP is trained on $400$ million image-text pairs (roughly $300\times$ greater than the size of ImageNet~\cite{deng2009imagenet}),
InstructGPT is trained on thousands of user-supplied and diverse prompts~\citep{ouyang2022training}, 
and LLaVA's instruction dataset contains over $100$ thousand image-text pairs~\citep{liu2023llava}.

These examples underscore the critical role of better designed and curated training data in further advancing the capabilities of LLMs. 
However, the heterogeneity, scale, and proprietary nature~\citep{bommasani2023_fm_transparency} of the training data for most of the currently best-performing LLMs significantly impede the progress in developing and training LLMs through curating better training data. 
To advance the research on data curation, we advocate for building towards rigorous data-centric benchmarks (\cref{sec:benchmarking}) on the foundation of existing efforts like DataComp~\citep{gadre2023datacomp}.

\paragraph{Data attribution.}
The training data is a ``source'' for the outputs generated by LLMs~\citep{keskar2019ctrl}. The ability to support source attribution and trace the generated outputs back to the specific training data is imperative for legal and safety purposes:
(i)~To respect the copyright/intellectual property rights, by correctly accrediting the creators of writings~\citep{eldan2023whos,Rahman2023copyright}, datasets~\citep{li2022untargetedwatermark,liu2023watermarkingdataset}, or code~\citep{lee2023codewatermark}.
(ii)~To mitigate the issue of problematic outputs of the LLMs (e.g., hateful, toxic, harmful messages~\citep{sap2019risk,Shelby2023taxonomyharm,Weidinger2022taxonomyrisk} or dangerous information~\citep{bommasani2022opportunities}), by identifying and removing the source.
Hence, we describe the promising directions for data attribution and removal (\cref{sec:attribution}).

\paragraph{Knowledge transfer.}
The costs of developing and deploying LLMs make it challenging to democratize the benefits of LLMs: GPT-4 costs over \$100 million to build~\citep{GPT4_100million} and is estimated to cost over \$21,000 a month for a small business to use for customer service support~\citep{chen2023frugalgpt}. 
Hence, a smaller model distilled from its larger counterparts for a specialized domain or task presents a cost-effective alternative~\citep{ jiang2023lion, taori2023stanford, yu2024dataset}.
The Zephyr $7$B beta outperforms the $70$B Llama $2$ in coding, math, and roleplay~\citep{tunstall2023zephyr} while MiniLLM matches the performance in instruction following of an LLM twice its parameter count~\citep{Gu2024minillm}.
These results open up promising avenues for transferring the knowledge of trained LLMs to compact and specialized models, and we discuss existing efforts and new opportunities where the outputs of a trained LLM are treated as (synthesized) data (\cref{sec:distillation}).

\paragraph{Inference contextualization with data.}
In contrast to standard ML models, LLMs have a unique capability of flexibly using data at inference to augment the outputs' factuality~\citep{wang2023shall} or quality~\citep{borgeaud2022improving}.
For example, an LLM can ``acquire'' a skill on the fly for a user's task via some user-provided examples~\citep{brown2020language}.
As another example, when queried, an LLM can search through a user-prepared datastore for relevant information as supplementary information for generating a response~\citep{lewis2020retrieval}. 
This capability enables the user to establish the right \emph{context} for the LLM at inference through the data (examples or datastore) and gives rise to an inference contextualization paradigm that can significantly streamline the applications of LLMs.
We elaborate on this paradigm w.r.t.~two prevalent technical frameworks and highlight how it can improve the personalization of LLMs~(\cref{sec:context}).

\section{Rigorous Data-centric Benchmarks} \label{sec:benchmarking}
There are increasing data-centric efforts on quantitatively understanding how the training data affects LLMs' performance via identifying and improving the scaling laws~\citep{deepseekllm_2024,hoffmann2022_compute_optimal,hu2024minicpm, kaplan2020scaling,sardana2023chinchillaoptimal}.
However, the datasets that train the state-of-the-art LLMs are often proprietary and closed-source while public datasets do not seem to achieve comparable scaling behavior to their proprietary counterparts~\citep{cherti2023reproducible}.
Moreover, even for public datasets like C4~\cite{raffel2020exploring} or LAION-2B~\cite{laion5b}, 
the critical factors underlying effective training datasets remain unclear.
Indeed, different training data compositions (i.e., proportions of different sources) can lead to vastly different properties of the trained CLIP models~\cite{nguyen2022quality} and language models \cite{anil2023palm,xie2023doremi}, while data filtering and pruning can sometimes even outperform the standard power-law scaling~\cite{abbas2023semdedup,sorscher2022beyond,toneva2018empirical}.
There are also promising results by sourcing for ``clean'' data~\citep{gunasekar2023textbooks_phi1} or low-perplexity data~\citep{marion2023less}.

These observations inspire several questions: What factors (besides the scale) are important to a training dataset~\citep{sachdeva2024train}? How do the data compositions affect the performance~\citep{xie2023doremi}? What is a principled methodology to reliably outperform the power-law scaling trends~\citep{sorscher2022beyond}?
While the above works provide excellent starting points, comprehensively addressing these questions requires a series of well-documented results and a systematic approach to identifying and quantitatively analyzing the key underlying factors of LLMs' performance. 
By building on the foundations in \citep{gadre2023datacomp,mazumder2023dataperf} which primarily target conventional ML, we advocate for rigorous data-centric benchmarks catering to LLMs' scale and complexity. 
We also identify directions (that leverage existing non-LLM-specific techniques) for designing effective data curation methods. 

\subsection{Research Directions: Benchmarks and Data Curation}
A cornerstone towards more efficient and effective LLM training powered by new data curation methods is rigorous and large-scale benchmarks for evaluation and results documentation.
The conventional ML benchmarking paradigm is completely flipped in these data-centric benchmarks~\citep{gadre2023datacomp} where the training code and computational budget are held constant so that participants innovate by proposing new training sets (e.g., new sources~\citep{gunasekar2023textbooks_phi1} or new filtering techniques~\citep{sachdeva2024train}).
We describe two specialized benchmarks, respectively, for designing training datasets and adapting to downstream domains and tasks,  
and further elaborate how they can be leveraged to design better methods for dataset design and curation.

\paragraph{Benchmarks for heterogeneous and large-scale pretraining data.}
Two key characteristics of the pretraining data of LLMs are \emph{heterogeneity} (e.g., multi-domain, multi-modality, multi-source) and the unprecedented \emph{scale}.
They induce not only an intricate interplay among the different domains, modalities, and sources but also a high complexity and cost of comprehensive evaluations~\citep{lee2023holistic,liang2023holistic}, thus making designing effective curation techniques challenging. 
Hence, instead of tackling the problem of curating pretraining data outright, we advocate for laying the foundations first by building benchmarks for heterogeneous and large-scale pretraining data based upon existing efforts such as DataComp \cite{gadre2023datacomp}.
DataComp is a benchmark for multimodal image-text dataset design for contrastive training of CLIP-like models.
Importantly, it spans several orders of magnitude in compute and data scale and includes the largest publicly available collection of over $6$ billion image-text pairs, making it a suitable testbed for testing hypotheses and drawing insights w.r.t.~the pretraining data for LLMs.
For example, one initial finding reports that changing how the training data is filtered led to significant improvements in CLIP-like models over OpenAI's original CLIP models~\citep{gadre2023datacomp}.
Compared to the few existing benchmarking efforts (often at a smaller data scale~\citep{datacentric}), DataComp is a more suitable starting point due to its scale and the promising initial findings.
Moreover, such benchmarking efforts can be complemented by the efforts on open-source pretraining datasets~\citep{Lozhkov2024fineweb-edu,Penedo2024fineweb,dolma2024}.

\paragraph{Benchmarks for adapting to downstream domains and tasks.}
Users usually want to apply LLMs to their downstream domains or tasks, motivating the investigation of how best to construct domain- or task-specific datasets to fine-tune an LLM pretrained on certain data.
For example, if we want to fine-tune a general-purpose LLM for medical tasks, does that general-purpose LLM need to have been pretrained on medical data (and if so, in what proportion), or does it suffice to fine-tune the LLM on a small amount of medical data?
As another example, to obtain an LLM for low-resource languages such as Southeast Asian~\cite{IMDA2023_SEAlanguages} or African languages~\cite{nguyen2023democratizing}, 
should we fine-tune an LLM pretrained on a mix of languages or one pretrained only on the target language?
Due to the specialized nature of these tasks, it is beneficial to explore more specialized adaptations of the existing benchmarking efforts. For multi-lingual adaptations (e.g., to adapt an LLM pretrained on English text to other languages), both Xtreme~\citep{hu2020xtremebenchmark} and TyDi~QA~\citep{clark2020tydi} benchmarks provide the resources for adequate evaluation and are thus suitable potential options.
For medical use cases, the CME~\citep{liu2023chinesemedbenchmark} and MedEval~\citep{he2023medevalbenchmark} benchmarks provide viable starting points.

\paragraph{Dataset design and curation.} 
The next step is developing methods for curating datasets for training LLMs and adapting them to downstream domains and tasks (i.e., fine-tuning).

For training general-purpose LLMs, the data needs to be diverse and spanning multiple distinct domains (e.g., books, Wikipedia, code, academic papers, etc.)~\cite{chowdhery2022palm} such that each domain is sufficiently well-represented in the training data to avoid overfitting~\citep{xie2023dsir}. 

The inter-domain and intra-domain curation processes have different requirements and thus separate considerations.
The inter-domain curation process should maximize heterogeneity, for instance, by incrementally selecting fine-grained domains~\citep{xie2023doremi} and adding in a new domain only if it adds to the heterogeneity of the pool of added domains. Statistical testing~\citep{gretton2012mmd,AISTATS21_wei2021sample} or distributional divergence~\citep{bendavid2010learning,ICML22_wu2022davinz} are principled methods to determine if a domain adds to the heterogeneity.
On the other hand, the intra-domain curation should maximize diversity~\citep{sachdeva2024train}, for instance, by integrating classic approaches such as determinantal point processes~\citep{Kulesza2012DPP} and coreset selection~\citep{sener2018active} with existing ML-based data valuation methods~\citep{Amiri2023_task_agnostic,IJCAI22_sim2022data}.

For adapting to downstream target domains or tasks, a core objective is to address the distribution shift between the target domain and the available training data; otherwise, the model learns irrelevant information about the target domain.
In this regard, a ``good'' data source has a high distributional similarity to the target domain. Hence, one can extend prior data valuation works in standard, unimodal ML settings~\citep{Amiri2023_task_agnostic, just2023lava} to efficiently handle multi-modal data at scale.
For selecting individual data points, prior works demonstrate the usefulness of influence scores~\citep{choe2024data, grosse2023studying, guo2021fastif, kwon2024datainf, xia2024less}.
\nocite{chen2023dataacquisition}

\subsection{Impact: Data-centric Open LLM Research}
With the benchmarking efforts and data curation methods, we hope to initiate a new brand of data-centric LLM research, welcoming openness and transparency.
While many efforts have been made to open-source the LLMs such as BLOOM~\citep{Scao2023bloom} and LLaMA~2~\citep{touvron2023llama}, most of the training data is held closed-source~\cite{bommasani2023_fm_transparency}.\footnote{A recent exception is the work of~\citet{olmo2024AI2}.} 
This new brand of data-centric open research can encourage more transparency in future research, which goes beyond the technological advancement itself but is also of great importance towards responsible adoptions of the technology and management of the ensuing socio-economical implications~\citep{bommasani2022opportunities,bommasani2023_fm_transparency}. For instance, the recently launched National AI Research Resource (NAIRR) by the U.S. National Science Foundation~\citep{NSF2024} lists open research (i.e., NAIRR open) as one of the four focus areas.

\section{Data Attribution} \label{sec:attribution}
For copyright/intellectual property rights considerations, data attribution is primarily motivated by the need for credit attribution.
For ensuring safe applications of LLMs, the goal of attribution is to trace (and then remove) the sources of potentially problematic outputs.
Notably, data attribution and unlearning are useful to both these use cases.

Since most of the training data for the popular LLMs is scraped from the Internet, it is almost inevitable that the training data contains certain copyrighted data (e.g., writing, code, or even entire datasets). 
Then, it is important to design techniques to mitigate potential copyright infringements, especially when the data owners or creators request takedowns.
This process involves first correctly identifying the source through data attribution and then removing it via unlearning~\citep{eldan2023whos}.
For sources that lead to problematic outputs by the LLMs, we first identify sources through attribution and then remove (the effects of) the sources through unlearning~\citep{si2023knowledge_ulLLMs}.
The challenge in the unlearning step is to ensure its effectiveness without compromising the performance of the LLM~\citep{chen2023unlearnllm}, incurring prohibitive costs from iterative retraining~\citep{si2023knowledge_ulLLMs} or needing additional training data~\citep{yao2023llmunlearn}.

\subsection{Research Directions: Data Attribution and Unlearning}
We describe data attribution followed by unlearning, which depends on data attribution.

\paragraph{Data attribution.}
We highlight two approaches where the first targets attribution to individual training data (i.e., more granular), and the second aims to identify a data source among several data sources (i.e., less granular).
For the first approach, attribution is by tracing the influence~\citep{koh2017understanding} or determining the value~\citep{ICML19_ghorbani2019data} of individual training data to LLMs.
While there have been successes in applying the influence function to attribute the prediction of an ML model to its training data~\citep{koh2019influence}, there remain challenges in extending it to LLMs.
The increasing complexity and size of model architectures significantly raise the computational cost~\citep{grosse2023studying} and deteriorate the influence scores' accuracy and utility~\citep{bae2022if,basu2021influence}.
Two promising approaches to address these computational challenges are efficient approximations~\citep{guo2021fastif,grosse2023studying}, and direct empirical estimators~\citep{guu2023simfluence,ilyas2022datamodels, pruthi2020estimating}.
Preliminary results demonstrate a computational speedup by reducing the original problem to a much smaller subproblem~\citep{guo2021fastif} or exploiting certain training structures~\citep{choe2024data,kwon2024datainf}.

The existing data valuation methods~\citep{ICML19_ghorbani2019data,Jia2019datavaluation,schoch2022csshapley,IJCAI22_sim2022data,Yoon2020datavaluationRL} can provide attribution by identifying the ``most valuable'' training data of a model (e.g., LLM). However, a similar scaling issue is encountered when applying these methods to LLMs, especially if they require multiple re-training of the LLM~\citep{schoch2023dataselectionllm}. 
Similarly, potential solutions include efficient approximations~\citep{VLDB19_jia2019efficient,schoch2023dataselectionllm} and training-free surrogates~\citep{just2023lava, ICLR23_nohyun2022data, ICML22_wu2022davinz} for designing scalable data valuation methods for LLMs.

For the second approach, source attribution differs from data attribution in being less granular and aiming to identify a data source instead of individual data.
This approach is particularly relevant in use cases involving copyrights or intellectual property rights, where the data source is the intellectual creator.
For source attribution, a natural idea is to adopt watermarking as a unique identifier for a piece of writing or design. For LLMs, watermarking techniques are used to identify or pinpoint the data sources that contribute most significantly to a given output~\citep{marra2018gans,yu2019attributing,yu2021artificial}.
Conceptually, a unique watermark is first assigned to each data source and then inserted into the training data from this source during training. Subsequently, given a generated output during inference, the most influential sources can be identified and correctly attributed by observing which of these watermarks are present in the output. Some specific types of watermarks include linguistic watermarks~\citep{kirchenbauer2023watermark,kuditipudi2023robust} and (non-linguistic) Unicode character-based watermarks~\citep{arXiv23_wang2023wasa}.

\paragraph{Unlearning of data.}
To remove (the effects of) certain identified training data (called target data), the set of unlearning techniques is suitable. 
The gold standard is to remove the target data and retrain the entire model from scratch on the remaining data, but it is prohibitively expensive for large models~\citep{SP15_cao2015towards,si2023knowledge_ulLLMs} and infeasible when regulations stipulate a short execution time~\citep{Graves2021}.
One alternative is to perform additional fine-tuning of the LLMs using only the remaining data to erase the effect of the target data~\citep{Mehta2022-zd, Neel2021-bv}. 
Another more directed approach is to leverage the knowledge of the target data to design cost-effective and efficient solutions, e.g., target data-oriented fine-tuning~\citep{yao2023llmunlearn} and in-context unlearning to ``mimic'' unlearning (the knowledge of specific tokens) via careful contextualization at inference time~\citep{pawelczyk2023incontextunlearn}.

\subsection{Impact: Safe and Responsible Deployment of LLM Technologies}
The ex-post data attribution and removal are useful for the safe and responsible deployment of LLMs by respecting the copyrights/intellectual property rights and mitigating problematic outputs.
These ex-post methods are complementary to possible ex-ante data-centric approaches (e.g., conditioning on certain types of data~\citep{keskar2019ctrl}) or other ex-post approaches (e.g., mitigation at decoding or inference time~\citep{krause2021gedi,liu2021dexperts}). Importantly, these methods target different stages of the LLM pipeline (i.e., before training, after training, and during inference) and collectively form ``multiple layers of defense'' against problematic outputs. Hence, we hope to inspire research towards ``multi-layered'' approaches for the safe and responsible deployments of LLM technologies.

\section{Knowledge Transfer} \label{sec:distillation}
Given the prohibitive costs of deploying full-fledged LLMs~\citep{chen2023frugalgpt, patterson2021carbon}, and that most users may not need such powerful general-purpose LLMs, the cost-effective adaptations of LLMs to users' specialized tasks are more appealing.
In many cases, the general-purpose LLM already has the necessary ``knowledge'' to perform the specialized task~\citep{li2023textbooks_phi15,xu2024survey_KD_LLM}, which can be transferred to a more compact and specialized model.
Knowledge transfer can be performed by first distilling the knowledge from the LLM as \emph{synthesized data}, then instilled into the specialized model by training it on the synthesized data. 
Since data synthesis is a niche setting arising from the generative capabilities of LLMs and its quality is key to effective knowledge transfers, we focus on data synthesis, specifically label and input syntheses.

\subsection{Research Directions: Cost-effective Data Synthesis}
We elaborate on label and input syntheses, focusing on the cost-effectiveness (i.e., the size/quantity and quality of the synthesized data).

\paragraph{Label synthesis.}
The simpler case is where the user starts with a large pool of unlabeled data (e.g., performing sentiment analysis for public comments) and requires the LLM to synthesize the labels.
This case resembles the setting of active learning~\citep{ICML17_gal2017deep} where the goal is cost-effectiveness (i.e., using a small number of synthesized labels to achieve a high learning performance).
The core idea is to select and annotate only the most ``useful'' data, which can be implemented via unsupervised data valuation techniques such as feature-based diversity~\citep{Amiri2023_task_agnostic, NeurIPS21_xu2021validation}, uncertainty modeling~\citep{lewis1994heterogeneous}, and optimized heuristics~\citep{bairi2015summarization}.
Additionally and different from the conventional unimodal settings, multi-modal classifiers like CLIP~\citep{ilharco2021openclip,radford2021learning} can be leveraged to perform cross-modal (e.g., image to text) or multi-modal (e.g., image-text to text) label synthesis. 

Moreover, the unique explanatory capabilities of LLMs can be exploited (i) to augment the synthesis with additional generated explanations and rationales~\citep{arXiv23_hsieh2023distilling}, and (ii) to be used, not as a ``label generator'' for direct label synthesis (as above), but as a labeled data ``selector''. Specifically, from a pool of labeled data (with labels possibly synthesized by an LLM), the LLM is asked to select the high-quality ones. It is useful when the original LLM cannot synthesize labels very accurately but is able to filter out the low-quality, noisy, or incorrect labels~\citep{sachdeva2024train}.

\paragraph{Input synthesis.}
The more challenging scenario arises when no initial data is available, not even unlabeled data, possibly because the specialized task is niche or less well-established and the user does not know what unlabeled data to collect.
We should fully utilize the generative capabilities of LLMs to synthesize coherent and diverse inputs~\citep{ding2024dataaug_llms}, such as via prompt engineering and fine-tuning procedures~\citep{li2021prefix} and sophisticated prompting techniques~\citep{naseh2024iteratively_multimodal}.
Then, the aforementioned label synthesis techniques can be applied, making label and input syntheses complementary to each other and suggesting it is possible to develop integrated treatments, such as jointly using existing unlabeled input and the generation of new input.
Notably, the $1.3$B phi-$1.5$ trained (almost) exclusively on synthesized data can outperform models $5\times$ larger~\citep{li2023textbooks_phi15} and the recently released Nemotron-4 family \citep{nvidia_nemotron340B_2024} further showcase the potential of synthesized data where over $98\%$ of data in their alignment process is synthesized.

\subsection{Impact: Democratization of the LLM Technologies}
The true testament to the impact of LLMs lies not in the streak of impressive metrics they score~\citep{arXiv23_openai2023gpt,srivastava2023beyondbigbench}
but rather in the concrete real-life successes~\citep[Impact Challenges]{Carbonell1992impactchallenges,Wagstaff2012mlmatters}. To do so requires the technology to be democratized and made accessible, not only through online API function calls but also in offline and resource-constrained environments, which is important to level the playing field for small organizations and individuals. 
We envision that the research directions of knowledge transfer can further widen the adoptions of LLMs (i) into different specialized domains including healthcare~\citep{savova2010mayo, yang2023large}, law~\citep{Law2024}, and education~\citep{Education2024}, (ii) at different scales, including consumer-grade hardware such as laptops~\citep{mlx2023} and smart-phones~\citep{AlpacaAndroid2023}, and (iii) in different scenarios where internet accessibility, data security and privacy concerns can present obstacles to users making use of the API function calls online~\citep{hao2022iron,liu2023llmsencryption}.

\section{Inference Contextualization with Data} \label{sec:context}

As described in the two examples in \cref{sec:introduction}, LLMs have the remarkable ability to utilize information ``in-context''~\citep{dong2023in-context-survey} where the context here is often in the form of a few example data points for demonstration or supplementary information~\citep{brown2020language}.
Such unique and unseen abilities present exciting use cases of data as an ``anchor'' to establish the right context at inference and enable the users to make certain specifications with flexibility and ease.

We illustrate such contextualization as follows:
(i) If a user prompts the LLM to generate a piece of writing while providing writings from Shakespeare, then the LLM's generated output can appear ``Shakespearean'' even though the LLM is not necessarily (extensively) trained on the writings from Shakespeare. (ii) If a user asks the LLM to solve a mathematical question while providing data containing similar questions and the reasoning steps, then the generated output can also contain reasoning steps, even though the LLM might not have been explicitly trained to do so.

\subsection{Research Directions: Data Selection for the Right Context}
For two technical frameworks that enable an LLM to utilize data at inference, namely retrieval-augmented generation (RAG) and in-context learning (ICL), we outline how LLMs utilize the data and then describe the corresponding research directions of data selection for contextualization.

\paragraph{Retrieval-augmented generation.}
RAG consists of two main components: the datastore and the retriever. The datastore is a collection of unstructured data (e.g., documents and their chunks), and structured data (e.g., as databases or knowledge graphs). 
Given a user query, (i) the retriever selects the most relevant and informative data from the datastore to (ii) contextualize the query for the LLM to generate an output~\citep{asai2024reliable_RAG}.
These two steps can be targeted as follows.

For (i), a more effective data selection (i.e., better relevance and informativeness) can be achieved by improving the indexing system of the datastore. Currently, the data (e.g., documents) in the datastore each has an indexing ``key'' (typically a vector in some embedding space~\citep{lewis2020retrieval,salton1975vector} containing some of the data's semantic meaning). However, for a Q\&A task, this indexing system can be ineffective for the retriever to identify the correct answer (i.e., data) to the question (i.e., query) since typically questions and answers have different semantic meanings. 
To improve its effectiveness, we can develop vector embeddings with built-in relevance in addition to the semantic meaning of data~\citep{formal2021splade,zamani2018neural} and pair them with the more classic approach of inverted index~\citep{zobel2006inverted} based on keywords and metadata.

For (ii), there are promising avenues of improvements targeting the different ways of how LLMs contextualize the query (i.e., utilizing the retrieved data) such as by improving the augmentation of the query with the retrieved data~\citep{Shi2024incontext, Yao2023react}, and designing a more effective ``fusion'' of the retrieved data and query~\citep{borgeaud2022improving,wang2023instructretro,wang2023shall}.

\paragraph{In-context learning.}
From a few user-provided demonstrations (i.e., data) in the query alone, LLMs can learn the hidden patterns and respond accordingly~\citep{dong2023in-context-survey}.
For example, to teach an LLM to solve mathematical questions, a user can query the LLM following this template: 

\texttt{Your goal is to solve math problems. Here are some examples: [EXAMPLES]. Now solve [QUESTION].}

The demonstration data, denoted as \texttt{[EXAMPLES]}, establishes the right context for the LLM.
Indeed the choice and quality of this demonstration data have a significant impact on the LLM's response quality (i.e., the correctness of the LLM's solution to \texttt{[QUESTION]})~\citep{ ACL22_lu2022fantastically, zhang2022active}.
Existing methods have shown the effectiveness of heuristics, including similarity~\citep{liu-etal-2022-makes}, uncertainty~\citep{arXiv23_diao2023active} and entropy~\citep{ACL22_lu2022fantastically}.
These results suggest opportunities for integrated frameworks with provable guarantees. 
For instance, optimization-based techniques have achieved preliminary successes in instruction optimization of LLMs (e.g., reinforcement learning~\citep{deng2022rlprompt}, Bayesian optimization~\citep{arXiv23_lin2023use} and evolutionary algorithms~\citep{guo2023connecting}), but have yet to be applied to optimize data selection in ICL.
Notably, a recent work utilizes neural bandits for the joint optimization of instructions and demonstration data~\citep{wu2024prompt}.

Note that RAG and ICL are not competing but rather complementary frameworks. With RAG, the user can leverage the size of the datastore for keeping more information while with ICL the user has an on-the-fly flexibility to direct specify the data with the query.

\subsection{Impact: Personalized Usages of LLMs}
Such contextualization (i.e., setting the context via specifying the data such as in RAG or ICL) has two hallmark practical benefits of being (i) simple and flexible via specifying the data and (ii) lightweight (i.e., no or minimal training/fine-tuning).
For a user, the data need not be static. For instance, a company using a RAG-powered Q\&A agent would, from time to time, update its product or service-related information. To ensure the Q\&A agent has updated information, updating the datastore would suffice. In contrast, updating the LLM via either training or fine-tuning can be time-consuming, costly, and technically complex, so personalization approaches (e.g., via RAG or ICL) that minimize or sidestep updating the LLM are more appealing in practice. 

Such features can simplify and make feasible the personalization of LLM technologies, which can have a significant impact on domains such as education~\citep{Alqahtani2023_benefits_education,gan2023LLMseducation,latif2023knowledge}
and healthcare~\citep{abbasian2023healthpersonalized,belyaeva2023hearlthpersonalized}.
LLMs-powered personalized curriculum designs can cater to the different needs of the students and educators can use LLMs to help prepare personalized feedback with significant time-saving benefits~\citep{Alqahtani2023LLMseducation}.
LLMs-based chatbots can provide timely personalized health assessments~\citep{Cascella2024_medicine}.

\section{Conclusion and Future Outlook}
This position paper has outlined a data-centric approach towards AI research with a focus on large language models~(LLMs).
We highlight the multi-faceted role of data in the different developmental (e.g., pretraining, fine-tuning) and inferential (e.g., data synthesis, inference contextualization) stages of LLMs.
In particular, we have identified four scenarios centered around data: rigorous data-centric benchmarks and data curation, data attribution, knowledge transfer, and inference contextualization with data. They each have unique challenges that require careful consideration, and present opportunities for innovation.

The impacts are described within each scenario for concreteness and clarity, but they are certainly \emph{not} restricted to each of the scenarios and can sometimes ``cross over''. For instance, while we have identified democratization of the LLM technologies as an impact of \cref{sec:distillation}, it is also applicable to \cref{sec:context}, which has highlighted the practical viability of personalized usages of LLMs.
Similarly, these scenarios (and the research directions therein) should not be viewed in isolation because there are indeed relationships and connections between the components. For instance, to mitigate problematic outputs by LLMs, a holistic treatment comprising both \emph{ex-ante} and \emph{ex-post} data-centric methods can perhaps be most effective (e.g., a more targeted data curation method from \cref{sec:benchmarking} paired with attribution and unlearning methods from \cref{sec:attribution}).

This initial exploration into a data-centric AI research paradigm in the age of LLMs is necessarily non-exhaustive and intended to catalyze broader discussions, stimulate further inquiry, and spark innovation that will expand the current limits of LLMs and, more broadly, AI, and build toward deployment of such technologies that promote greater democratization.

\section{Limitations and Impact Statement} \label{sec:limitations}
This section organizes the limitations and alternative viewpoints following the same organization as the main paper.

\paragraph{On~\cref{sec:benchmarking}.}
One limitation of the outlined research directions is that these directions do not specifically account for the interplay between different steps (e.g., pretraining data and fine-tuning data) or between model (e.g., architecture and size) and data~\citep{hoffmann2022_compute_optimal,sardana2023chinchillaoptimal}.
It is an appealing next step to develop integrated pipelines covering data curation methods for different steps and jointly leverage model-centric and data-centric insights.

\paragraph{On~\cref{sec:attribution}.}
One specific limitation of using fine-tuning to achieve unlearning is that its effectiveness is limited if there are only a small number of fine-tuning iterations due to a short stipulated execution time or a small fine-tuning dataset~\citep{Golatkar2020-el}. As a result, after unlearning via fine-tuning, the model might still contain traces of the ``deleted'' target data. This limitation can be mitigated by adopting more directed unlearning techniques such as those described in \cref{sec:attribution}.

We differentiate our described data-centric watermarking approaches (for data attribution) from existing model-oriented watermarking methods~\citep{huang2024optimalstatswatermark,kuditipudi2023robust,zhao2023provable,Zhao2023invisiblewatermark} (for determining whether a given output is generated by LLMs or a specific LLM).
Additionally, we differentiate our described unlearning approaches (for removing or erasing certain target data) from knowledge unlearning~\citep{si2023knowledge_ulLLMs}, whose goal is to forget an abstract definition of knowledge~\citep{chen2023unlearnllm,jang2023kul,wang2023kga_unlearning}.

\paragraph{On~\cref{sec:distillation}.}
A key requirement for effective knowledge transfer is that the general-purpose LLM has the ``necessary'' knowledge.
This requirement is not always satisfied as there are areas where even the most advanced LLMs are lacking (e.g., reasoning and planning~\citep{Dziri2023_compose_reasoning,valmeekam2023_reasoning}).
Nevertheless, there are many areas and use cases for which existing open-sourced LLMs are very capable~\citep{olmo2024AI2,nvidia_nemotron340B_2024} and can be used for knowledge transfer, and data synthesis in general.
Furthermore, even if the LLM is not able to perform label synthesis optimally, it can still be useful for filtering out low-quality labels and leaving the good labels for training, as in ``impossible distillation''~\citep{jung2023impossible}.

Another possible limitation in practice is due to the possible legal restrictions of how/whether existing proprietary and closed-source LLMs can be used, especially for commercial purposes.\footnote{\href{https://openai.com/policies/terms-of-use}{Terms of use, OpenAI.}}\footnote{ \href{https://console.anthropic.com/legal/terms}{Terms of Service, Anthropic.}}\footnote{ \href{https://ai.google.dev/terms}{Generative AI APIs Additional Terms of Service, Google.}}
Nevertheless, there are more efforts underway to open-source and democratize LLM technologies~\citep{vicuna2023,liu2023llava,taori2023stanford,touvron2023llama}. For instance, \citet{olmo2024AI2} completely open-sourced their LLM, including the pretraining data, model architecture, and trained weights, and the entire training logs, under the Apache-2.0 license, permitting a ``free'' use of this trained model, such as for knowledge transfer.
As another example, \citet{nvidia_nemotron340B_2024} released the Nemotron-4 family and their entire synthetic data generation pipeline under the NVIDIA Open Model License,\footnote{\href{https://developer.download.nvidia.com/licenses/nvidia-open-model-license-agreement-june-2024.pdf}{NVIDIA Open Model License Agreement}.} allowing the distribution, modification, and use of the models and its outputs.

\paragraph{On~\cref{sec:context}.}
One limitation of the inference contextualization is that it is difficult to design foolproof techniques or guarantees due to the complexity and the intricate black-box internal working mechanism of LLMs. It may require additional future investigation to understand and then leverage the mechanism of LLMs to design techniques with provable guarantees.
Our position is to highlight a practically simple and technically viable approach for personalizing LLMs, as well as the promising research directions and techniques.

\subsection*{Impact Statement}
This position paper presents a data-centric viewpoint towards AI research with a focus on LLMs, outlining specific scenarios for future research and highlighting the respective impacts therein.

\section*{Acknowledgements}
This research is supported by the National Research Foundation Singapore and the Singapore Ministry of Digital Development and Innovation, National AI Group under the AI Visiting Professorship Programme (award number AIVP-$2024$-$001$). 
This research/project is supported by the National Research Foundation, Singapore under its AI Singapore Programme (AISG Award No: AISG$2$-PhD/$2021$-$08$-$017$[T]). 
This research/project is supported by the National Research Foundation, Singapore under its AI Singapore Programme (AISG Award No: AISG$2$-PhD/$2023$-$01$-$039$J). 
This research is part of the programme DesCartes and is supported by the National Research Foundation, Prime Minister’s Office, Singapore under its Campus for Research Excellence and Technological Enterprise (CREATE) programme. 
This research is supported by the National Research Foundation (NRF), Prime Minister's Office, Singapore under its Campus for Research Excellence and Technological Enterprise (CREATE) programme. The Mens, Manus, and Machina (M3S) is an interdisciplinary research group (IRG) of the Singapore MIT Alliance for Research and Technology (SMART) centre.  

\bibliographystyle{plainnat}
\bibliography{references}  

\end{document}